\documentclass[a4paper]{llncs}

\usepackage{amssymb}
\usepackage{amsmath}
\usepackage{graphicx}
\usepackage{todonotes}
\usepackage{url}
\usepackage{caption}
\urldef{\mailsa}\path|{per-arne.andersen, morten.goodwin, ole.granmo}@uia.no|
\newcommand{\keywords}[1]{\par\addvspace\baselineskip
\noindent\keywordname\enspace\ignorespaces#1}

\begin{document}

\mainmatter  

\title{Towards a Deep Reinforcement Learning Approach for Tower Line Wars}

\titlerunning{Towards a Deep Reinforcement Learning Approach for Tower Line Wars}

%
%
\author{Per-Arne Andersen%
\and Morten Goodwin\and Ole-Christoffer Granmo}
\authorrunning{Towards a Deep Reinforcement Learning Approach for Tower Line Wars}

\institute{University of Agder, Grimstad, Norway \\
\url{cair-internal@uia.no}}

%
%

\toctitle{Towards a Deep Reinforcement Learning Approach for Tower Line Wars}

\maketitle

\begin{abstract}
There have been numerous breakthroughs with reinforcement learning in the recent years, perhaps most notably on Deep Reinforcement Learning successfully playing and winning relatively advanced computer games. There is undoubtedly an anticipation that Deep Reinforcement Learning will play a major role when the first AI masters the complicated game plays needed to beat a professional Real-Time Strategy game player. For this to be possible, there needs to be a game environment that targets and fosters AI research, and specifically Deep Reinforcement Learning. Some game environments already exist, however, these are either overly simplistic such as Atari 2600 or complex such as Starcraft II from Blizzard Entertainment.
 
We propose a game environment in between Atari 2600 and Starcraft II, particularly targeting Deep Reinforcement Learning algorithm research. The environment is a variant of Tower Line Wars from Warcraft III, Blizzard Entertainment. Further, as a proof of concept that the environment can harbor Deep Reinforcement algorithms, we propose and apply a Deep Q-Reinforcement architecture. The architecture simplifies the state space so that it is applicable to Q-learning, and in turn improves performance compared to current state-of-the-art methods. Our experiments show that the proposed architecture can learn to play the environment well, and score 33\% better than standard Deep Q-learning --- which in turn proves the usefulness of the game environment.



\end{abstract}

\section{Introduction}
\label{cap:introduction}
Despite many advances in AI for games, no universal reinforcement learning algorithm can be applied to Real-Time Strategy Games (RTS) without data manipulation or customization. This includes traditional games such as Warcraft III, Starcraft II, and Tower Line Wars. Reinforcement Learning (RL) has been applied to simpler games such as games for the Atari 2600 platform but has to the best of our knowledge not successfully been applied to RTS games. Further, existing game environments that target AI research are either overly simplistic such as Atari 2600 or complex such as Starcraft II.

Reinforcement Learning has had tremendous progress in recent years in learning to control agents from high-dimensional sensory inputs like vision. In simple environments, this has been proven to work well \cite{mnih-atari-2013}, but are still an issue for complex environments with large state and action spaces \cite{DBLP:journals/corr/MirowskiPVSBBDG16}. In games where the objective is easily observable, there is a short distance between action and reward which fuels the learning. This is because the consequence of any action is quickly observed, and then easily learned. When the objective is more complicated the game objectives still need to be mapped to the reward function, but it becomes far less trivial. For the Atari 2600 game Ms. Pac-Man this was solved through a hybrid reward architecture that transforms the objective to a low-dimensional representation \cite{arXiv:1706.04208}. Similarly, the OpenAI's bot is able to beat world's top professionals at 1v1 in DotA 2. It uses reinforcement learning while it plays against itself, learning to predict the opponent moves. 

Real-Time Strategy Games, including Warcraft III, is a genre of games much more comparable to the complexity of real-world environments. It has a sparse state space with many different sensory inputs that any game playing algorithm must be able to master in order to perform well within the environment. Due to the complexity and because many action sequences are required to constitute a reward, standard reinforcement learning techniques including Q-learning are not able to master the games successfully. 

This paper introduces a two-player version of the popular Tower Line Wars modification from the game Warcraft III. We refer to this variant as Deep Line Wars. Note that Tower Line Wars is not an RTS game, but has many similar elements such as time-delayed objectives, resource management, offensive, and defensive strategy planning. To prove that the environment is working we, inspired by recent advances from van Seijen et al. \cite{arXiv:1706.04208}, apply a method of separating the abstract reward function of the environment into smaller rewards. This approach uses a Deep Q-Network using a Convolutional Neural Network to map actions to states and can play the game successfully and perform better than standard Deep Q-learning by 33\%. 

Rest of the paper is organized as follows:
We first investigate recent discoveries in Deep RL in section \ref{cap:related_work}. We then briefly outline how Q-Learning works and how we interpret Bellman's equation for utilizing Neural Networks as a function approximator in section \ref{cap:q_learning}. We present our contribution in section \ref{cap:deep_line_wars} and present a comparison of other game environments that are widely used in reinforcement learning. We introduce a variant of Deep Q-Learning in section \ref{cap:deep_q_reward_network} and present a comparison to other RL models used in state-of-the-art research. Finally we show results in section \ref{cap:experiments}, define a roadmap of future work in section \ref{cap:future_work} and conclude our work in section \ref{cap:conclusion}

\section{Related Work}
\label{cap:related_work}
\begin{figure}
\centering
\includegraphics[width=0.99\linewidth]{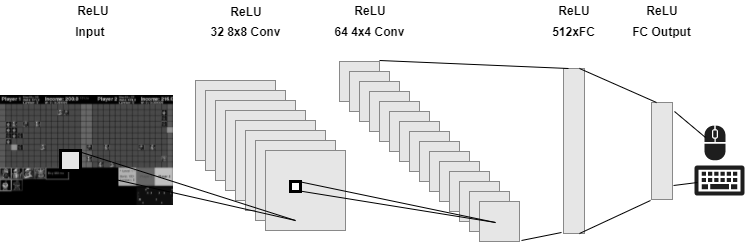}
\caption{Deep Q-Learning architecture}
\label{fig:q_learning_arch}
\end{figure}

There have been several breakthroughs related to reinforcement learning performance in recent years \cite{gosavi2009reinforcement}. Q-Learning together with Deep Learning was a game-changing moment, and has had tremendous success in many single agent environments on the Atari 2600 platform \cite{mnih-atari-2013}. Deep Q-Learning as proposed by Mnih et al. \cite{mnih-atari-2013} as shown in Figure \ref{fig:q_learning_arch} used a neural network as a function approximator and outperformed human expertise in over half of the games \cite{mnih-atari-2013}.

Hasselt et al. proposed Double DQN, which reduced the overestimation of action values in the Deep Q-Network \cite{DBLP:journals/corr/HasseltGS15}. This led to improvements in some of the games on the Atari platform. 

Wang et al. then proposed a dueling architecture of DQN which introduced estimation of the value function and advantage function \cite{DBLP:journals/corr/WangFL15}. These two functions were then combined to obtain the Q-Value. Dueling DQN were implemented with the previous work of van Hasselt et al. \cite{DBLP:journals/corr/WangFL15}. 

Harm van Seijen et al. recently published an algorithm called Hybrid Reward Architecture (HRA) which is a divide and conquer method where several agents estimate a reward and a Q-value for each state \cite{arXiv:1706.04208}. The algorithm performed above human expertise in Ms. Pac-Man, which is considered one of the hardest games in the Atari 2600 collection and is currently state-of-the-art in the reinforcement learning domain \cite{arXiv:1706.04208}.
The drawback of this algorithm is that generalization of Minh et al. approach is lost due to a huge number of separate agents that have domain-specific sensory input.

There have been few attempts at using Deep Q-Learning on advanced simulators specifically made for machine-learning. It is probable that this is because there are very few environments created for this purpose.


\section{Q-Learning}
\label{cap:q_learning}
Reinforcement learning can be considered hybrid between supervised and unsupervised learning. We implement what we call an agent that acts in our environment. This agent is placed in the unknown environment where it tries to maximize the environmental reward \cite{Sutton98a}.

Markov Decision Process (MDP) is a mathematical method of modeling decision-making within an environment. We often use this method when utilizing model-based RL algorithms. In Q-Learning, we do not try to model the MDP. Instead, we try to learn the optimal policy by estimating the action-value function $Q^*(s, a)$, yielding maximum expected reward in state s executing action a. The optimal policy can then be fosund by 

\begin{equation}
\pi(s) = argmax_aQ^*(s,a)
\end{equation}

This is derived from \textit{Bellman's Equation}, because we can consider $U(s) = max_aQ(s,a)$, the Utility function to be true. This gives us the ability to derive following update-rule equation from Bellman's work:

\begin{equation}
Q(s,a) \leftarrow Q(s,a) + \underbrace{\alpha}_\text{Learning Rate} \Bigg( \underbrace{R(s)}_\text{Reward} + \underbrace{\gamma}_\text{Discount} \underbrace{max_{a^{'}} Q(s^{'},a^{'})}_\text{New Estimate} - \underbrace{Q(s,a)}_\text{Old Estimate}  \Bigg)
\end{equation}

This is an iterative process of propagating back the estimated Q-value for each discrete time-step in the environment. It is guaranteed to converge towards the optimal action-value function, $Q_i \rightarrow Q^*$ as i $\rightarrow \infty$ \cite{Sutton98a,mnih-atari-2013}.
At the most basic level, Q-Learning utilize a table for storing $(s,a,r,s^{'})$ pairs. But we can instead use a non-linear function approximation in order to approximate $Q(s,a;\theta)$. $\theta$ describes tunable parameters for approximator. Artificial Neural Networks (ANN) are a popular function approximator, but training using ANN is relatively unstable.
We define the loss function as following.

\begin{equation}
\label{eq:policy_iteration}
L(\theta_{i}) = E\Big[(r + \gamma max_{a^{'}}Q(s^{'},a^{'}; \theta_{i}) - Q(s,a;\theta_{i}))^2\Big]
\end{equation}

As we can see, this equation uses Bellman equation to calculate the loss for the gradient descent.
To combat training instability, we use \textit{Experience Replay}. This is a memory module which stores memories from experienced states and draws a uniform distribution of experiences to train the network \cite{mnih-atari-2013}. This is what we call a \textit{Deep Q-Network} and are as described in its most primitive form. See related work for recent advancements in DQN.

\section{Deep Line Wars}
\label{cap:deep_line_wars}
For a player to play RTS games well, he typically needs to master high difficulty strategies. Most RTS strategies incorporate 
\begin{itemize}
    \item Build strategies, 
    \item Economy management, 
    \item Defense evaluation, and
    \item Offense evaluation.
\end{itemize}
These objectives are easy to master when separated but become hard to perfect when together. Starcraft II is one of the most popular RTS games, but due to its complexity, it is not expected that an AI-based system can beat this game anytime soon. At the very least, state-of-the-art Deep Q-Learning is not directly applicable. Blizzard entertainment and Google DeepMind has collaborated on an interface to the Starcraft II game. \cite{starcraftiiforums,vinyals_2016}. Starcraft II is for many researchers considered the next big goal in AI research. Warcraft III is relatable to Starcraft II as they are the same genre and have near identical game mechanics.

Current state-of-the-art algorithms struggle to learn objectives in the state-space because the action-space is too abstract. \cite{DBLP:journals/corr/LillicrapHPHETS15}. State and action spaces define the range of possible configurations a game board can have. Existing DQN models use pixel data as input and objectively maps state to action \cite{mnih-atari-2013}. This works when the game objective is closely linked to an action, such as controlling a paddle in Breakout, where the correct action is quickly rewarded, and a wrong action quickly punished. This is not possible in RTS games. If the objective is to win the game, an action will only be rewarded or punished after minutes or even hours of gameplay. Furthermore, gameplay would consist of thousands of actions and only combined will they result in a reward or punishment.

\begin{figure}[!htp]
\centering
\includegraphics[width=0.90\linewidth]{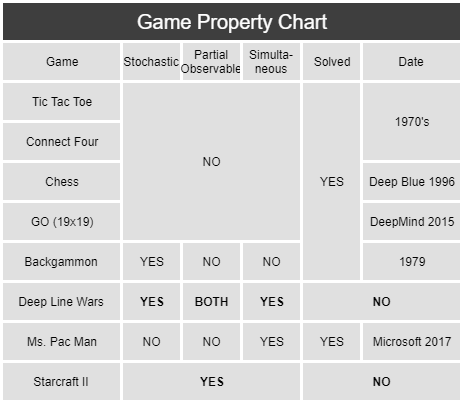}
\caption{Properties of selected game environments}
\label{fig:property_chart}
\end{figure}

Collected data in Figure \ref{fig:property_chart} argues that games that have been solved by current state-of-the-art is usually non-stochastic and is fully observable. Also, current AI prefers environments which are not simultaneous, meaning they can be paused between each state transition. This makes sense because hardware still limits advances in AI.

By doing rough estimations of the state-space in-game environments from Figure \ref{fig:property_chart}, it is clear that state-of-the-art has done a big leap in recent years. With the most recent contribution being Ms. Pac-Man \cite{arXiv:1706.04208}. However, by computing the state-space of a regular Starcraft II map only taking unit compositions into account, the state space can be calculated to be $(128x128)^{400} = 16384^{400} = 10^{1685}$ \cite{uriarte_ontañón_2014}.

\begin{figure}
\centering
\includegraphics[width=0.99\linewidth]{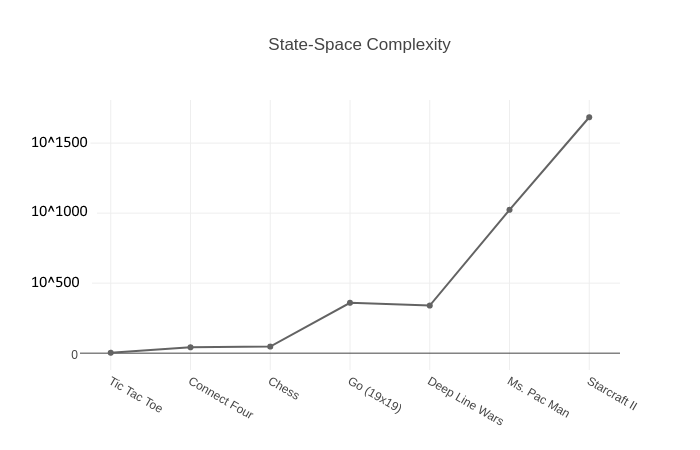}
\caption{State-space complexity of selected game environments}
\label{fig:state_space}
\end{figure}

The predicament is that the difference in complexity between Ms. Pac-Man and Starcraft II is tremendous. Figure \ref{fig:state_space} illustrates a relative and subjective comparison between state-complexity in relevant game environments. State-space complexity describes approximately how many different game configurations a game can have. It is based on map size, unit position, and unit actions. The comparison is a bit arbitrary because the games are complex in different manners. However, there is no doubt that the distance between Ms. Pac-Man, perhaps the most advanced computer game mastered so far, and Starcraft II is colossal. To advance AI solutions towards Starcraft II, we argue that there is a need for several new game environments that exceed the complexity of existing games and challenge researches on multi-agent issues closely related to Starcraft II \cite{DBLP:journals/corr/abs-1207-4708}.
We, therefore, introduce Deep Line Wars as a two-player variant of Tower Line Wars. Deep Line Wars is a game simulator aimed at filling the gap between Atari 2600 and Starcraft II. It features the most important aspects of an RTS game.

\begin{figure}[!ht]
\centering
\includegraphics[height=6.2cm]{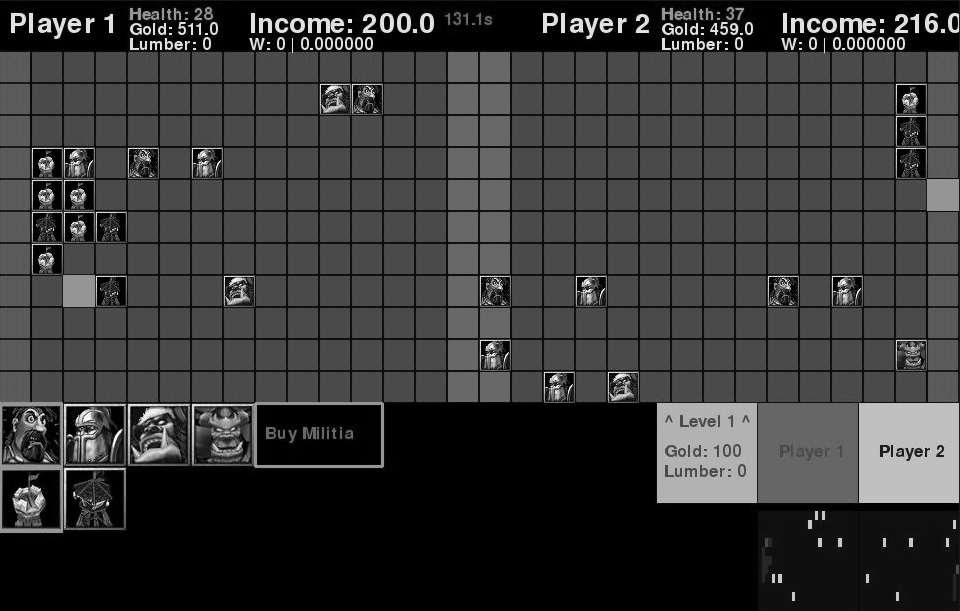}
\caption{Graphical Interface of Deep Line Wars}
\label{fig:deep_line_wars}
\end{figure}

The objective of this game is as seen in Figure \ref{fig:deep_line_wars} to invade the opposing player with units until all health is consumed. The opposing player’s health is reduced for each friendly unit that enters the red area of the map. A unit spawns at a random location on the red line of the controlling player’s side and automatically walks towards the enemy base. To protect your base against units, the player can build towers which shoot projectiles at enemy units. When an enemy unit dies, a fair percentage of the unit value is given to the player. When a player sends a unit, the income variable is increased by a defined percentage of the unit value. Players gold are increased at regular intervals determined in the configuration files.
To master Deep Line Wars, the player must learn following skill-set: 
\begin{itemize}
    \item offensive strategies of spawning units,
    \item defending against the opposing player's invasions, and
    \item maintain a healthy balance between offensive and defensive in order to maximize income
\end{itemize}
and is guaranteed a victory if mastered better than the opposing player.

\begin{figure}[!htp]
  \centering
  \includegraphics[height=2.3cm]{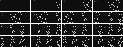}
  \caption{State abstraction using gray-scale heat-maps}
  \label{fig:heatmaps}
\end{figure}

Because the game is specifically targeted towards machine learning, the game-state is defined as a multi-dimensional matrix.
\begin{figure}
\centering
\includegraphics[width=0.90\linewidth]{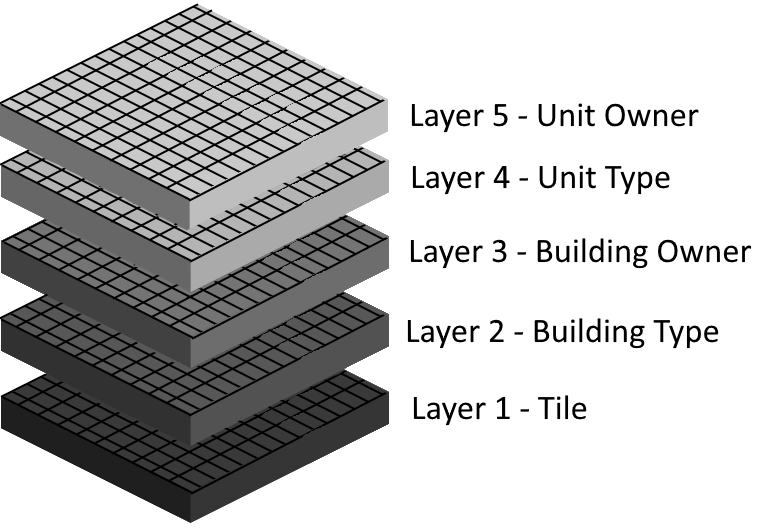}
\caption{Game-state representation}
\label{fig:game_architecture}
\end{figure}
Figure \ref{fig:game_architecture} represents a 5x30x11 state-space that contains all relevant board information at current time-step. It is therefore easy to cherry-pick required state-information when using it in algorithms. 
Deep Line Wars also features possibilities of making an abstract representation of the state-space, seen in Figure \ref{fig:heatmaps}. This is a heat-map that represent the state (Figure \ref{fig:game_architecture}) as a lower-dimensional state-space. Heat-maps also allows the developer to remove noise that causes the model to diverge from the optimal policy, see Formula \ref{eq:policy_iteration}.

We need to reduce the complexity of the state-space to speed up training. Using heat-maps made it possible to encode the five-dimensional state information into three dimensions. These dimensions are RGB values that we can find in imaging. Figure \ref{fig:heatmaps} show how the state is seen from the perspective of player 1 using gray-scale heatmaps. We define
\begin{itemize}
    \item red pixels as friendly buildings, 
    \item green pixels as enemy units, and
    \item teal pixels as the mouse cursor.
\end{itemize}
We also included an option to reduce the state-space to a one-dimensional matrix using gray-scale imaging. Each of the above features is then represented by a value between 0 and 1. We do this because Convolutional Neural Networks are computational demanding, and by reducing input dimensionality, we can speed up training. \cite{mnih-atari-2013} We do not down-scale images because the environment is only 30x11 pixels large. The state cannot be described fully by these heat-maps as there are economics, health, and income that must be interpreted separately. This is solved by having a 1-dimensional vectorized representation of the data, that can be fed into the model.

\section{DeepQRewardNetwork}
\label{cap:deep_q_reward_network}
The main contribution in this paper is the game environment presented in Section \ref{cap:deep_line_wars}. A key element is to show that the game environment is working properly and we, therefore, introduce a learning algorithm trying to play the game. This is in no way meant as a perfect solver for Deep Line Wars, but rather as a proof of concept that learning algorithms can be applied in the Deep Line Wars environment. 
In our solution we consider the environment as a MDP having state set S, action set A, and a reward function set R. Each of the weighted reward functions derives from a specific agent within the MDP and defines the absolute reward of the environment $R_{env}$ with following equation:

\begin{equation}
R_{env}(s,a) = \sum\limits_{i=1}^n w_{i}R_{i}(s,a)
\end{equation}

Where $R_{env}(s,a)$ is the weighted sum $w_{i}$ of reward function(s) $R_{i}(s,a)$. 
The proposed algorithm model is a method of dividing the ultimate problem into separate smaller problems which can be trivialized with certain kinds of generic algorithms.

\begin{figure}
\centering
\includegraphics[width=0.99\linewidth]{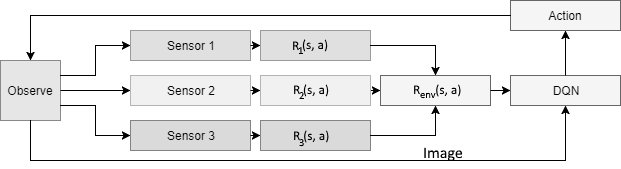}
\caption{Separation of the reward function}
\label{fig:reward_algorithm}
\end{figure}

When reward for the observed state is calculated, we calculate the Q-value of $Q(s, a)$ utilizing $R_{env}$ by using a variant of DQN.

\section{Experiments}
\label{cap:experiments}
We conducted experiments with several deep learning algorithms in order to benchmark current state-of-the-art put up against a multi-agent, multi-sensory environment. The experiments were conducted in Deep Line Wars, a multi-agent, multi-sensory environment. All algorithms were benchmarked with identical game parameters. 

\begin{figure}[!htp]
\centering
\includegraphics[width=0.99\linewidth]{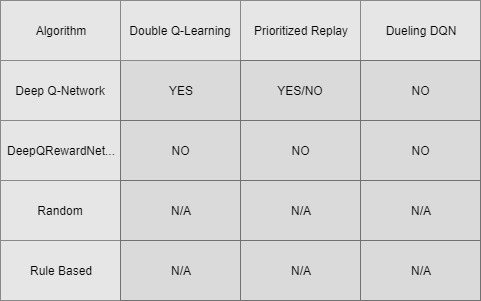}
\caption{Property matrix of tested algorithms}
\label{fig:experiments_matrix}
\end{figure}

We tested \textit{DeepQNetwork}, a state-of-the-art DQN from Mnih et al\cite{mnih-atari-2013}, \textit{DeepQRewardNetwork}, rule-based, and random behaviour. Each of the algorithms was tested with several configurations, seen in Figure \ref{fig:experiments_matrix}. We did not expect any of these algorithms to beat the rule-based challenge due to the difficulty of the AI. The extended execution graph algorithm (see Section \ref{cap:future_work}) was not part of the test bed because it was not able to compete with any of the simpler DQN algorithms without guided mouse management. 

Tests were done using Intel I7-4770k, 64GB RAM and NVIDIA Geforce GTX 1080TI. Each of the algorithms was trained/executed for 1500 episodes. Each episode is considered to be a game that either of the players wins, or the 600 seconds time limit is reached. DQN had a discount-factor of 0.99, learning rate of 0.001 and batch-size of 32.


\begin{figure}[!htp]
\centering
\includegraphics[width=0.99\linewidth]{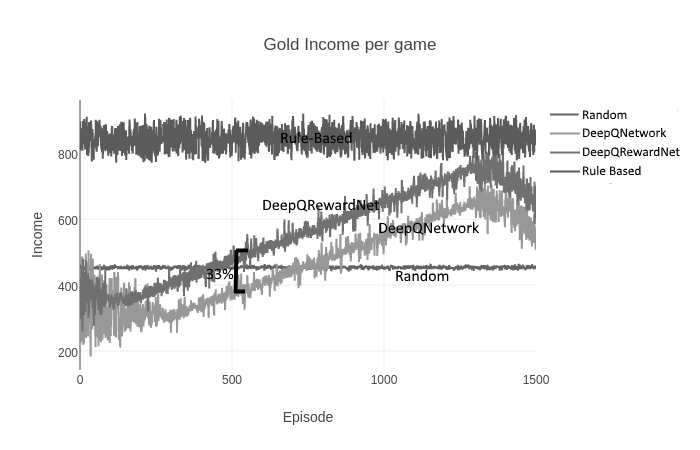}
\caption{Income after each episode}
\label{fig:results_avg_income}
\end{figure}

Throughout the learning process, we can see that DeepQNetwork and DeepQRewardNetwork learn to perform resource management correctly. Figure \ref{fig:results_avg_income} illustrates income throughout learning from 1500 episodes. The random player is presented as an aggregated average of 1500 games, but the remaining algorithms are only single instances. It is not practical to perform more than a single run of the Deep Learning algorithms because it takes several minutes per episode to finish which sums up to a huge learning time.

Figure \ref{fig:results_avg_income} shows that the proposed algorithms outperform random behavior after relatively few episodes. DeepQRewardNetwork performs approximately 33\% better than DeepQNetwork. We believe that this is because the reward function $R(s, a)$ is better defined and therefore easier to learn the optimal policy in a shorter period of time. These results show that DeepQRewardNetwork converges towards the optimal policy better, but as seen in Figure \ref{fig:results_avg_income} diverges after approximately 1300 games. The reason for the divergence is that experience replay does not correctly batch important memories to the model. This causes the model to train on unimportant memories and diverges the model. This is considered a part of future work and is addressed more thoroughly in section \ref{cap:future_work}. The rule-based algorithm can be regarded as an average player and can be compared to human level in this game environment. 

\begin{figure}[!htp]
\centering
\includegraphics[width=0.99\linewidth]{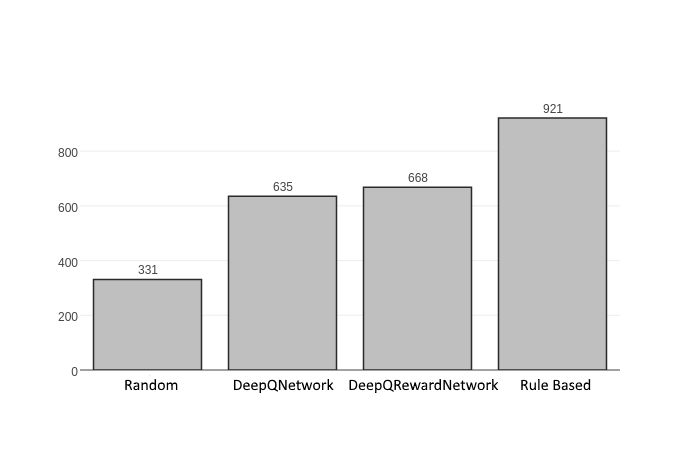}
\caption{Victory distribution of tested algorithms}
\label{fig:results_win_distrib}
\end{figure}

Figure \ref{fig:results_win_distrib} shows that DeepQNetwork and DeepQRewardNetwork have about 63-67\% win ratio throughout the learning process. Compared to the rule-based AI it does not qualify to be near mastering the game, but we can see that it outperforms random behavior in the game environment.

\section{Future Work}
\label{cap:future_work}
This paper introduced a new learning environment for reinforcement learning and applied state-of-the-art Deep-Q Learning to the problem. Some initial results showed progress towards an AI that could beat a rule-based AI. There are still several challenges that must be addressed for an unsupervised AI to learn complex environments like Line Tower Wars. Mouse input based games are difficult to map to an abstract state representation, because there are a huge number of sequenced mouse clicks that are required, to correctly act in the game. DQN cannot at current state handle long sequences of actions and must be guided in-order to succeed. Finding a solution to this problem without guiding is thought to be the biggest blocker for these types of environments, and will be the focus for future work.

DeepQNetwork and DeepQRewardNetwork had issues with divergence after approximately 1300 episodes. This is because our experience replay algorithm did not take into account that the majority of experiences are bad. It could not successfully prioritize the important memories. As future work, we propose to instead use prioritized experience replay from Schaul et al. \cite{DBLP:journals/corr/SchaulQAS15}.

\begin{figure}[!htp]
\centering
\includegraphics[height=4.2cm]{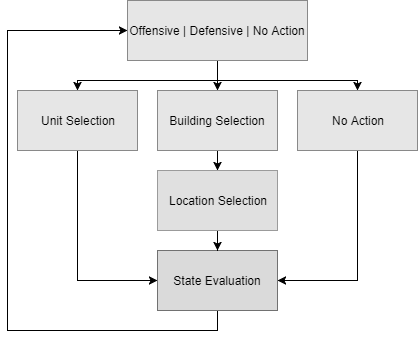}
\caption{Divide \& Conquer Execution graph}
\label{fig:execution_graph}
\end{figure}
Figure \ref{fig:reward_algorithm} show that different sensors separate the reward from the environment to obtain a more precise reward bound to an action. In our research, we developed an algorithm that utilizes different models based on which state the player has. Figure \ref{fig:execution_graph} show the general idea, where the state is categorized into three different types \textit{Offensive}, \textit{Defensive}, and \textit{No Action}. This state is evaluated by a Convolutional Neural Network and outputs a one-hot vector that signal which state the player is currently in. Each of the blocks in Figure \ref{fig:execution_graph} then represents a form of state-modeling that is determined by the programmer. Our initial tests did not yield any promising results, but according to the Bellman equations, it is a qualified way of evaluating the state and successfully perform learning, on an iterative basis. 

\section{Conclusion}
\label{cap:conclusion}
Deep Line Wars is a simple but yet advanced Real-Time (strategy) game simulator, which attempts to fill the gap between Atari 2600 and Starcraft II. DQN shows promising initial results but is far from perfect in current state-of-the-art. An attempt in making abstractions in the reward signal yielded some improved performance, but at the cost of a  more generalized solution. Because of the enormous state-space, DQN cannot compete with simple rule-based algorithms. We believe that this is caused by specifically the mouse input which requires some understanding of the state to perform well.
This also causes the algorithm to overestimate some actions, specifically the offensive actions, because the algorithm is not able to correctly build defensive without getting negative rewards.
It is imperative that a solution of the mouse input actions are found before DQN can perform better. A potential approach could be using the StarCraft II API to get additional training data, including mouse sequences \cite{2017arXiv170804782V}.


\end{document}